\documentclass[11pt,a4paper]{article}
\usepackage[hyperref]{emnlp2018}
\usepackage{times}
\usepackage{latexsym}
\usepackage{url}
\usepackage{bm}

\usepackage{cleveref}
\crefformat{section}{\S#2#1#3} 
\crefformat{subsection}{\S#2#1#3}
\crefformat{subsubsection}{\S#2#1#3}

\usepackage{algorithm}
\usepackage{algorithmic}
\usepackage{enumerate}
\usepackage{amsmath} 
\usepackage{amssymb}  

\usepackage{booktabs}
\usepackage{multirow}
\usepackage{float}
\usepackage{graphicx}
\usepackage{footnote}
\usepackage{tablefootnote}
\usepackage{subcaption}

\usepackage{latexsym}
\usepackage{cleveref}
\crefformat{section}{\S#2#1#3} 
\crefformat{subsection}{\S#2#1#3}
\crefformat{subsubsection}{\S#2#1#3}
\usepackage{algorithm}
\usepackage{algorithmic}
\usepackage{enumerate}
\usepackage{amsmath} 
\usepackage{amssymb}  
\usepackage{booktabs}
\usepackage{multirow}
\usepackage{float}
\usepackage{graphicx}
\usepackage{footnote}
\usepackage{tablefootnote}
\usepackage{subcaption,siunitx,booktabs}
\usepackage{graphics}
\usepackage{stfloats}

\aclfinalcopy 

\newcommand*{\affaddr}[1]{#1} 
\newcommand*{\affmark}[1][*]{\textsuperscript{#1}}
\newcommand*{\email}[1]{\texttt{#1}}

\title{Adaptive Semi-supervised Learning for Cross-domain \\ Sentiment Classification}

\author{Ruidan He\affmark[\dag\ddag], Wee Sun Lee\affmark[\dag], Hwee Tou Ng\affmark[\dag], \and Daniel Dahlmeier\affmark[\ddag]\\
\affaddr{\affmark[\dag]Department of Computer Science, National University of Singapore}\\
\affaddr{\affmark[\ddag]SAP Innovation Center Singapore}\\
\email{\affmark[\dag]\{ruidanhe,leews,nght\}@comp.nus.edu.sg}\\
\email{\affmark[\ddag]d.dahlmeier@sap.com}%
}

\date{}

\begin{document}
\maketitle
\begin{abstract}
We consider the cross-domain sentiment classification problem, where a sentiment classifier is to be learned from a source domain and to be generalized to a target domain. 
Our approach explicitly minimizes the distance between the source and the target instances in an embedded feature space. With the difference between source and target minimized, we then exploit additional information from the target domain by consolidating the idea of semi-supervised learning, for which, we jointly employ two regularizations -- entropy minimization and self-ensemble bootstrapping -- to incorporate the unlabeled target data for classifier refinement. Our experimental results demonstrate that the proposed approach can better leverage unlabeled data from the target domain and achieve substantial improvements over baseline methods in various experimental settings. 
\end{abstract}

\section{Introduction}
In practice, it is often difficult and costly to annotate sufficient training data for diverse application domains on-the-fly. We may have sufficient labeled data in an existing domain (called the source domain), but very few or no labeled data in a new domain (called the target domain).  This issue has motivated research on cross-domain sentiment classification, where knowledge in the source domain is transferred to the target domain in order to alleviate the required labeling effort.

One key challenge of domain adaptation is that data in the source and target domains are drawn from different distributions. Thus, adaptation performance will decline with an increase in distribution difference. Specifically, in sentiment analysis, reviews of different products have different vocabulary. For instance, restaurants reviews would contain opinion words such as ``tender'', ``tasty'', or ``undercooked'' and movie reviews would contain ``thrilling'', ``horrific'', or ``hilarious''. The intersection between these two sets of opinion words could be small which makes domain adaptation difficult. 

Several techniques have been proposed for addressing the problem of domain shifting. The aim is to bridge the source and target domains by learning domain-invariant feature representations so that a classifier trained on a source domain can be adapted to another target domain. In cross-domain sentiment classification, many works ~\cite{blitzer:07,pan:10,zhou:15,wu:16,Yu:16} utilize a key intuition that domain-specific features could be aligned with the help of domain-invariant features (pivot features). For instance, ``hilarious'' and ``tasty'' could be aligned as both of them are relevant to ``good''. 

Despite their promising results, these works share two major limitations. First, they highly depend on the heuristic selection of pivot features, which may be sensitive to different applications. Thus the learned new representations may not effectively reduce the domain difference. Furthermore, these works only utilize the unlabeled target data for representation learning while the sentiment classifier was solely trained on the source domain. There have not been many studies on exploiting unlabeled target data for refining the classifier, even though it may contain beneficial information. How to effectively leverage unlabeled target data still remains an important challenge for domain adaptation.

In this work, we argue that the information from unlabeled target data is beneficial for domain adaptation and we propose a novel \textbf{D}omain \textbf{A}daptive \textbf{S}emi-supervised learning framework (DAS) to better exploit it. Our main intuition is to treat the problem as a semi-supervised learning task by considering target instances as unlabeled data, assuming the domain distance can be effectively reduced through domain-invariant representation learning. Specifically, the proposed approach jointly performs feature adaptation and semi-supervised learning in a multi-task learning setting. For feature adaptation, it explicitly minimizes the distance between the encoded representations of the two domains. On this basis, two semi-supervised regularizations -- entropy minimization and self-ensemble bootstrapping -- are jointly employed to exploit unlabeled target data for classifier refinement.

We evaluate our method rigorously under multiple experimental settings by taking label distribution and corpus size into consideration. The results show that our model is able to obtain significant improvements over strong baselines. We also demonstrate through a series of analysis that the proposed method benefits greatly from incorporating unlabeled target data via semi-supervised learning, which is consistent with our motivation. Our datasets and source code can be obtained from \url{https://github.com/ruidan/DAS}.

\section{Related Work}
\textbf{Domain Adaptation}:
The majority of feature adaptation methods for sentiment analysis rely on a key intuition that even though certain opinion words are completely distinct for each domain, they can be aligned if they have high correlation with some domain-invariant opinion words (pivot words) such as ``excellent'' or ``terrible''. Blitzer et al. (\citeyear{blitzer:07}) proposed a method based on structural correspondence learning (SCL), which uses pivot feature prediction to induce a projected feature space that works well for both the source and the target domains. The pivot words are selected in a way to cover common domain-invariant opinion words. Subsequent research aims to better align the domain-specific words~\cite{pan:10,he:11,wu:16} such that the domain discrepancy could be reduced. More recently, Yu and Jiang (\citeyear{Yu:16}) borrow the idea of pivot feature prediction from SCL and extend it to a neural network-based solution with auxiliary tasks. In their experiment, substantial improvement over SCL has been observed due to the use of real-valued word embeddings. 
Unsupervised representation learning with deep neural networks (DNN) such as denoising autoencoders has also been explored for feature adaptation~\cite{glorot:11,chen:12,Yang:14}. It has been shown that DNNs could learn transferable representations that disentangle the underlying factors of variation behind data samples. 

Although the aforementioned methods aim to reduce the domain discrepancy, they do not explicitly minimize the distance between distributions, and some of them highly rely on the selection of pivot features. In our method, we formally construct an objective for this purpose. Similar ideas have been explored in many computer vision problems, where the representations of the underlying domains are encouraged to be similar through explicit objectives~\cite{tzeng:14,ganin:15,long:15,zhuang:15,Long:17} such as maximum mean discrepancy (MMD)~\cite{gretton:12}. In NLP tasks, Li et al. (\citeyear{Li:17}) and Chen et al. (\citeyear{Chen:17}) both proposed using adversarial training framework for reducing domain difference. In their model, a sub-network is added as a domain discriminator while deep features are learned to confuse the discriminator. The feature adaptation component in our model shares similar intuition with MMD and adversary training. We will show a detailed comparison with them in our experiments.
\smallskip

\noindent\textbf{Semi-supervised Learning}: 
We attempt to treat domain adaptation as a semi-supervised learning task by considering the target instances as unlabeled data. Some efforts have been initiated on transfer learning from unlabeled data~\cite{dai:07,jiang:07,Wu:09}. In our model, we reduce the domain discrepancy by feature adaptation, and thereafter adopt semi-supervised learning techniques to learn from unlabeled data.
Primarily motivated by~\cite{grandvalet:04} and~\cite{Laine:17}, we employed entropy minimization and self-ensemble bootstrapping as regularizations to incorporate unlabeled data. Our experimental results show that both methods are effective when jointly trained with the feature adaptation objective, which confirms to our motivation.

\section{Model Description}
 
\subsection{Notations and Model Overview}
We conduct most of our experiments under an unsupervised domain adaptation setting, where we have no labeled data from the target domain. Consider two sets $D_s$ and $D_t$. $D_s = \{\mathbf{x}_i^{(s)}, \mathbf{y}_i^{(s)}\} |_{i=1}^{n_s}$ is from the source domain with $n_s$ labeled examples, where $\mathbf{y}_i \in \mathbb{R}^C$ is a one-hot vector representation of sentiment label and $C$ denotes the number of classes. $D_t = \{\mathbf{x}_i^{(t)}\} |_{i=1}^{n_t}$ is from the target domain with $n_t$ unlabeled examples. $N=n_s+n_t$ denotes the total number of training documents including both labeled and unlabeled\footnote{Note that unlabeled source examples can also be included for training. In that case, $N=n_s+n_t+n_{s^\prime}$ where $n_{s^\prime}$ denotes the number of unlabeled source examples. This corresponds to our experimental setting 2. For simplicity, we only consider $n_s$ and $n_t$ in our description.}. We aim to learn a sentiment classifier from $D_s$ and $D_t$ such that the classifier would work well on the target domain. We also present some results under a setting where we assume that a small number of labeled target examples are available (see Figure~\ref{label_target}).

For the proposed model, we denote $G$ parameterized by $\theta_{g}$ as a neural-based feature encoder that maps documents from both domains to a shared feature space, and $\mathcal{F}$ parameterized by $\theta_{f}$ as a fully connected layer with softmax activation serving as the sentiment classifier. We aim to learn feature representations that are domain-invariant and at the same time discriminative on both domains, thus we simultaneously consider three factors in our objective: (1) minimize the classification error on the labeled source examples; (2) minimize the domain discrepancy; and (3) leverage unlabeled data via semi-supervised learning. 

Suppose we already have the encoded features of documents $\{\bm{\xi}_i^{(s,t)} = G(\mathbf{x}_i^{(s,t)}; \theta_g)\}|_{i=1}^N$ (see Section \ref{implement}),  the objective function for purpose (1) is thus the cross entropy loss on the labeled source examples 
\begin{equation}\label{crossentropy_obj}
L = -\frac{1}{n_s}\sum_{i=1}^{n_s}\sum_{j=1}^C \mathbf{y}_i^{(s)}(j) \log \tilde{\mathbf{y}}^{(s)}_i(j)
\end{equation}
where $\tilde{\mathbf{y}}_i^{(s)} = \mathcal{F}(\bm{\xi}^{(s)}_i; \theta_f)$ denotes the predicted label distribution. In the following subsections, we will explain how to perform feature adaptation and domain adaptive semi-supervised learning in details for purpose (2) and (3) respectively.

\subsection{Feature Adaptation}
Unlike prior works~\cite{blitzer:07, Yu:16}, our method does not attempt to align domain-specific words through pivot words. In our preliminary experiments, we found that word embeddings pre-trained on a large corpus are able to adequately capture this information. 
As we will later show in our experiments, even without adaptation, a naive neural network classifier with pre-trained word embeddings can already achieve reasonably good results. 

We attempt to explicitly minimize the distance between the source and target feature representations ($\{\bm{\xi}_i^{(s)}\}|_{i=1}^{n_s}$ and $\{\bm{\xi}_i^{(t)}\}_{i=1}^{n_t}$). A few methods from literature can be applied such as Maximum Mean Discrepancy (MMD)~\cite{gretton:12} or adversary training~\cite{Li:17,Chen:17}. The main idea of MMD is to estimate the distance between two distributions as the distance between sample means of the projected embeddings in Hilbert space. MMD is implicitly computed through a characteristic kernel, which is used to ensure that the sample mean is injective, leading to the MMD being zero if and only if the distributions are identical. In our implementation, we skip the mapping procedure induced by a characteristic kernel for simplifying the computation and learning. We simply estimate the distribution distance as the distance between the sample means in the current embedding space. Although this approximation cannot preserve all statistical features of the underlying distributions, we find it performs comparably to MMD on our problem. The following equations formally describe the feature adaptation loss $\mathcal{J}$:
\begin{align}
&\mathcal{J} = KL(\mathbf{g}_s || \mathbf{g}_t) + KL(\mathbf{g}_t || \mathbf{g}_s)\label{feature_adaptation_obj} \\
&\mathbf{g}_s^{\prime} = \frac{1}{n_s}\sum_{i=1}^{n_s}\bm{\xi}_i^{(s)} \text{,} 
\quad \quad \mathbf{g}_s = \frac{\mathbf{g}_s^{\prime}}{\|\mathbf{g}_s^{\prime}\|_1} \\
&\mathbf{g}_t^{\prime} = \frac{1}{n_t}\sum_{i=1}^{n_t}\bm{\xi}_i^{(t)} \text{,} 
\quad \quad \mathbf{g}_t = \frac{\mathbf{g}_t^{\prime}}{\|\mathbf{g}_t^{\prime}\|_1} 
\end{align}
$L_1$ normalization is applied on the mean representations $\mathbf{g}_s^{\prime}$ and $\mathbf{g}_t^{\prime}$, rescaling the vectors such that all entries sum to 1. We adopt a symmetric version of KL divergence~\cite{zhuang:15} as the distance function. Given two distribution vectors $\mathbf{\emph{P}}, \mathbf{\emph{Q}} \in \mathbb{R}^{k}$, $KL(\mathbf{\emph{P}} || \mathbf{\emph{Q}}) = \sum_{i=1}^{k} \mathbf{\emph{P}}(i)\log(\frac{\mathbf{\emph{P}}(i)}{\mathbf{\emph{Q}}(i)})$. 

\subsection{Domain Adaptive Semi-supervised Learning (DAS)}
We attempt to exploit the information in target data through semi-supervised learning objectives, which are jointly trained with $L$ and $\mathcal{J}$. Normally, to incorporate target data, we can minimize the cross entropy loss between the true label distributions $\mathbf{y}_i^{(t)}$ and the predicted label distributions $\tilde{\mathbf{y}}_i^{(t)} = \mathcal{F}(\bm{\xi}^{(t)}_i; \theta_f)$ over target samples. The challenge here is that $\mathbf{y}_i^{(t)}$ is unknown, and thus we attempt to estimate it via semi-supervised learning. We use entropy minimization and bootstrapping for this purpose. We will later show in our experiments that both methods are effective, and jointly employing them overall yields the best results.
\smallskip

\noindent\textbf{Entropy Minimization:} 
In this method, $\mathbf{y}_i^{(t)}$ is estimated as the predicted label distribution $\tilde{\mathbf{y}}_i^{(t)}$, which is a function of $\theta_g$ and $\theta_f$. The loss can thus be written as 
\begin{equation}\label{entropy_minimization_obj}
\Gamma = -\frac{1}{n_{t}}\sum_{i=1}^{n_t}\sum_{j=1}^{C} \tilde{\mathbf{y}}^{(t)}_i(j) \log \tilde{\mathbf{y}}^{(t)}_i(j)
\end{equation}
Assume the domain discrepancy can be effectively reduced through feature adaptation, by minimizing the entropy penalty, training of the classifier is influenced by the unlabeled target data and will generally maximize the margins between the target examples and the decision boundaries, increasing the prediction confidence on the target domain.
\smallskip

\noindent\textbf{Self-ensemble Bootstrapping:}
Another way to estimate $\mathbf{y}_i^{(t)}$ corresponds to bootstrapping. The idea is to estimate the unknown labels as the predictions of the model learned from the previous round of training. Bootstrapping has been explored for domain adaptation in previous works~\cite{jiang:07,Wu:09}. However, in their methods, domain discrepancy was not explicitly minimized via feature adaptation. Applying bootstrapping or other semi-supervised learning techniques in this case may worsen the results as the classifier can perform quite bad on the target data.  

\begin{algorithm}[t]
\caption{Pseudocode for training DAS }
\begin{algorithmic} 
\REQUIRE $D_s$, $D_t$, $G$, $\mathcal{F}$
\REQUIRE $\alpha$ = ensembling momentum, $0\leq\alpha<1$
\REQUIRE $w(t)$ = weight ramp-up function
\STATE $\mathbf{Z} \leftarrow 0_{[N \times C]}$
\STATE $\tilde{\mathbf{z}} \leftarrow 0_{[N \times C]}$
\FOR{$t \in [1, \text{\emph{max-epochs}}]$}
	\FOR{each minibatch $B^{(s)}$, $B^{(t)}$, $B^{(u)}$ in \\
     \qquad	\qquad $D_s$, $D_t$, $\{x_i^{(s,t)}\}|_{i=1}^{N}$}
    \STATE compute loss $L$ on $[\mathbf{x}_{i \in B^{(s)}}, \mathbf{y}_{i \in B^{(s)}}]$
    \STATE compute loss $\mathcal{J}$ on $[\mathbf{x}_{i \in B^{(s)}}, \mathbf{x}_{j \in B^{(t)}}]$
    \STATE compute loss $\Gamma$ on $\mathbf{x}_{i \in B^{(t)}}$
    \STATE compute loss $\Omega$ on $[\mathbf{x}_{i \in B^{(u)}}, \tilde{\mathbf{z}}_{i \in B^{(u)}}]$
    \STATE $\textrm{\emph{overall-loss}} \leftarrow L+\lambda_1 \mathcal{J} + \lambda_2 \Gamma + w(t) \Omega$ \\ 
    update network parameters
    \ENDFOR
    \STATE $\mathbf{Z}^{\prime}_i \leftarrow  \mathcal{F}(G(\mathbf{x}_i))$, for $i \in N$
    \STATE $\mathbf{Z} \leftarrow \alpha \mathbf{Z}+(1-\alpha)\mathbf{Z}^{\prime}$
    \STATE $\tilde{\mathbf{z}} \leftarrow \text{\emph{one-hot-vectors}}(\mathbf{Z})$
\ENDFOR
\end{algorithmic}\label{algo}
\end{algorithm}

Inspired by the ensembling method proposed in~\cite{Laine:17}, we estimate $\mathbf{y}_i^{(t)}$ by forming ensemble predictions of labels during training, using the outputs on different training epochs. The loss is formulated as follows:
\begin{equation}
\Omega = -\frac{1}{N}\sum_{i=1}^{N}\sum_{j=1}^{C} \tilde{\mathbf{z}}^{(s,t)}_i(j) \log \tilde{\mathbf{y}}^{(s,t)}_i(j)
\end{equation}
where $\tilde{\mathbf{z}}$ denotes the estimated labels computed on the ensemble predictions from different epochs. The loss is applied on all documents. It serves for bootstrapping on the unlabeled target data, and it also serves as a regularization that encourages the network predictions to be consistent in different training epochs. 
$\Omega$ is jointly trained with $L$, $\mathcal{J}$, and $\Gamma$.
Algorithm \ref{algo} illustrates the overall training process of the proposed domain adaptive semi-supervised learning (DAS) framework.

In Algorithm \ref{algo}, $\lambda_1$, $\lambda_2$, and $w(t)$ are weights to balance the effects of $\mathcal{J}$, $\Gamma$, and $\Omega$ respectively. $\lambda_1$ and $\lambda_2$ are constant hyper-parameters. We set $w(t)=\exp[-5(1-\frac{t}{\text{\emph{max-epochs}}})^2]\lambda_3$ as a Gaussian curve to ramp up the weight from $0$ to $\lambda_3$. This is to ensure the ramp-up of the bootstrapping loss component is slow enough in the beginning of the training. After each training epoch, we compute $\mathbf{Z}^{\prime}_i$ which denotes the predictions made by the network in current epoch, and then the ensemble prediction $\mathbf{Z}_i$ is updated as a weighted average of the outputs from previous epochs and the current epoch, with recent epochs having larger weight. For generating estimated labels $\tilde{\mathbf{z}}_i$, $\mathbf{Z}_i$ is converted to a one-hot vector where the entry with the maximum value is set to one and other entries are set to zeros. 
The self-ensemble bootstrapping is a generalized version of bootstrappings that only use the outputs from the previous round of training~\cite{jiang:07,Wu:09}. The ensemble prediction is likely to be closer to the correct, unknown labels of the target data.

\section{Experiments}

\renewcommand{\arraystretch}{1.1}
\begin{table}[t]
\begin{subtable}{0.5\textwidth}
\centering
\resizebox{0.6\columnwidth}{!}{
\begin{tabular}{l|lrrr|r}
\toprule
\textbf{Domain} && \textbf{\#Pos} & \textbf{\#Neg} & \textbf{\#Neu} & \textbf{Total} \\\hline
Book & Set 1 & 2000 & 2000 & 2000 & 6000 \\
& Set 2 & 4824 & 513 & 663 & 6000\\\hline
Electronics & Set 1 & 2000 & 2000 & 2000 & 6000\\
& Set 2 & 4817 & 694 & 489 & 6000\\\hline
Beauty & Set 1 & 2000 & 2000 & 2000 & 6000\\
& Set 2 & 4709 & 616 & 675 & 6000\\\hline
Music & Set 1 & 2000 & 2000 & 2000 & 6000\\
& Set 2 & 4441 & 785 & 774 & 6000\\
\bottomrule
\end{tabular}
}
\caption{Small-scale datasets}\label{small-scale-data}
\end{subtable}
\medskip

\begin{subtable}{0.5\textwidth}
\centering
\resizebox{0.6\columnwidth}{!}{
\begin{tabular}{l|rrr|r}
\toprule
\textbf{Domain} & \textbf{\#Pos} & \textbf{\#Neg} & \textbf{\#Neu} & \textbf{Total} \\\hline
IMDB & 55,242 & 11,735 & 17,942 & 84,919 \\
Yelp & 155,625 & 29,597 & 45,941 & 231,163\\
Cell Phone & 148,657 & 24,343 & 21,439 & 194,439\\
Baby & 126,525 & 17,012 & 17,255 & 160,792 \\
\bottomrule
\end{tabular}
}
\caption{Large-scale datasets}\label{large-scale-data}
\end{subtable}
\caption{Summary of datasets.}
\end{table}


\subsection{CNN Encoder Implementation}\label{implement}
We have left the feature encoder $G$ unspecified, for which, a few options can be considered. In our implementation, we adopt a one-layer CNN structure from previous works~\cite{Kim:14,Yu:16}, as it has been demonstrated to work well for sentiment classification tasks. 
Given a review document $\mathbf{x} = (x_1, x_2, ..., x_n)$ consisting of $n$ words, we begin by associating each word with a continuous word embedding~\cite{Mikolov:13} $\mathbf{e}_x$ from an embedding matrix $\mathbf{E} \in \mathbb{R}^{V \times d}$, where $V$ is the vocabulary size and $d$ is the embedding dimension. $E$ is jointly updated with other network parameters during training. Given a window of dense word embeddings $\mathbf{e}_{x_1}, \mathbf{e}_{x_2}, ... , \mathbf{e}_{x_l}$, the convolution layer first concatenates these vectors to form a vector $\mathbf{\hat{x}}$ of length $ld$ and then the output vector is computed by Equation (\ref{conv}):
\begin{equation}
Conv(\mathbf{\hat{x}}) = f(\mathbf{W} \cdot \mathbf{\hat{x}} + \mathbf{b})
\label{conv}
\end{equation}
$\theta_{g} = \{\mathbf{W}$, $\mathbf{b}\}$ is the parameter set of the encoder $G$ and is shared across all windows of the sequence. $f$ is an element-wise non-linear activation function. The convolution operation can capture local contextual dependencies of the input sequence and the extracted feature vectors are similar to $n$-grams. After the convolution operation is applied to the whole sequence, we obtain a list of hidden vectors $\mathbf{H} = (\mathbf{h}_1, \mathbf{h}_2, ..., \mathbf{h}_n)$. A max-over-time pooling layer is applied to obtain the final vector representation $\bm{\xi}$ of the input document.

\subsection{Datasets and Experimental Settings}
Existing benchmark datasets such as the Amazon benchmark~\cite{blitzer:07} typically remove reviews with neutral labels in both domains. This is problematic as the label information of the target domain is not accessible in an unsupervised domain adaptation setting. Furthermore, removing neutral instances may bias the dataset favorably for max-margin-based algorithms like ours, since the resulting dataset has all uncertain labels removed, leaving only high confidence examples. Therefore, we construct new datasets by ourselves. The results on the original Amazon benchmark is qualitatively similar, and we present them in Appendix~\ref{append_amazon} for completeness since most of previous works reported results on it.
\smallskip

\noindent\textbf{Small-scale datasets:} Our new dataset was derived from the large-scale Amazon datasets\footnote{http://jmcauley.ucsd.edu/data/amazon/} released by McAuley et al. (\citeyear{Mcauley:15}). It contains four domains\footnote{The original reviews were rated on a 5-point scale. We label them with rating $<3$, $>3$, and $=3$ as negative, positive, and neutral respectively.}: Book (BK), Electronics (E), Beauty (BT), and Music (M).  Each domain contains two datasets. Set 1 contains 6000 instances with exactly balanced class labels, and set 2 contains 6000 instances that are randomly sampled from the large dataset, preserving the original label distribution, which we believe better reflects the label distribution in real life. The examples in these two sets do not overlap. Detailed statistics of the generated datasets are given in Table~\ref{small-scale-data}. 

In all our experiments on the small-scale datasets, we use set 1 of the source domain as the only source with sentiment label information during training, and we evaluate the trained model on set 1 of the target domain. Since we cannot control the label distribution of unlabeled data during training, we consider two different settings: 
\smallskip

\noindent{\emph{Setting (1):}} Only set 1 of the target domain is used as the unlabeled set. This tells us how the method performs in a condition when the target domain has a close-to-balanced label distribution. As we also evaluate on set 1 of the target domain, this is also considered as a transductive setting.
\smallskip

\noindent{\emph{Setting (2):}} Set 2 from both the source and target domains are used as unlabeled sets. Since set 2 is directly sampled from millions of reviews, it better reflects real-life sentiment distribution.
\smallskip

\noindent\textbf{Large-scale datasets:} We further conduct experiments on four much larger datasets: IMDB\footnote{IMDB is rated on a 10-point scale, and we label reviews with rating $<5$, $>6$, and $=5/6$ as negative, positive, and neutral respectively.} (I), Yelp2014 (Y), Cell Phone (C), and Baby (B). IMDB and Yelp2014 were previously used in~\cite{tang:15,yang:17}. Cell phone and Baby are from the large-scale Amazon dataset~\cite{Mcauley:15,He:16}.  Detailed statistics are summarized in Table~\ref{large-scale-data}.
We keep all reviews in the original datasets and consider a transductive setting where all target examples are used for both training (without label information) and evaluation. We perform sampling to balance the classes of labeled source data in each minibatch $B^{(s)}$ during training.

\subsection{Selection of Development Set}
Ideally, the development set should be drawn from the same distribution as the test set. However, under the unsupervised domain adaptation setting, we do not have any labeled target data at training phase which could be used as development set. In all of our experiments, for each pair of domains, we instead sample 1000 examples from the training set of the source domain as development set. We train the network for a fixed number of epochs, and the model with the minimum classification error on this development set is saved for evaluation. This approach works well on most of the problems since the target domain is supposed to behave like the source domain if the domain difference is effectively reduced.

Another problem is how to select the values for hyper-parameters. If we tune $\lambda_1$ and $\lambda_2$ directly on the development set from the source domain, most likely both of them will be set to 0, as unlabeled target data is not helpful for improving in-domain accuracy of the source domain. Other neural network models also have the same problem for hyper-parameter tuning. Therefore, our strategy is to use the development set from the target domain to optimize $\lambda_1$ and $\lambda_2$ for one problem (e.g., we only do this on E$\rightarrow$BK), and fix their values on the other problems. This setting assumes that we have at least two labeled domains such that we can optimize the hyper-parameters, and then we fix them for other new unlabeled domains to transfer to. 

\begin{figure*}[t!]
    \centering
    \begin{subfigure}[b]{0.8\textwidth}
        \centering
        \includegraphics[width=\textwidth]{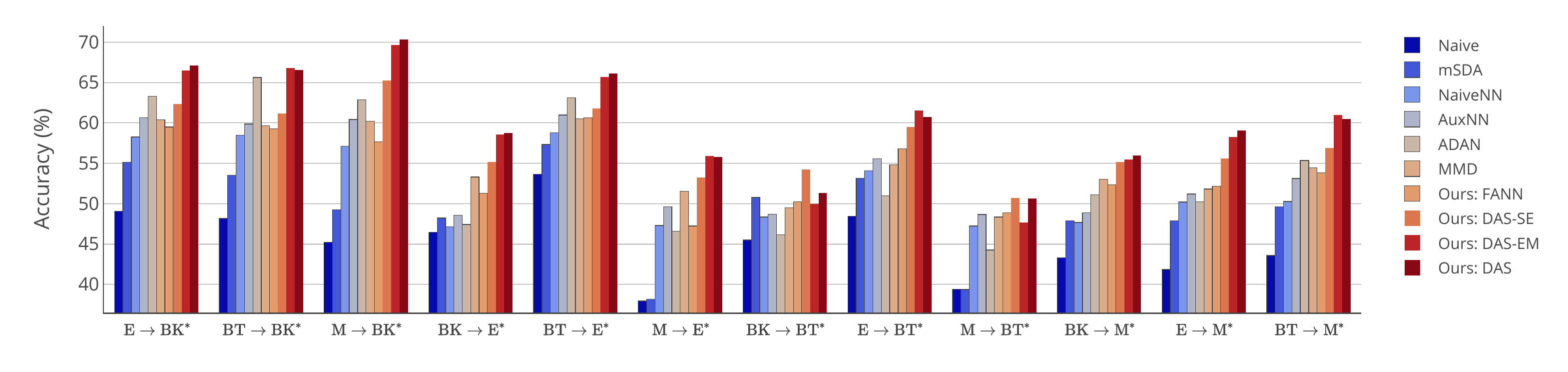}
        \caption{Accuracy on the small-scale dataset under setting 1.}
    \end{subfigure}
    
    \begin{subfigure}[b]{0.8\textwidth}
        \centering
        \includegraphics[width=\textwidth]{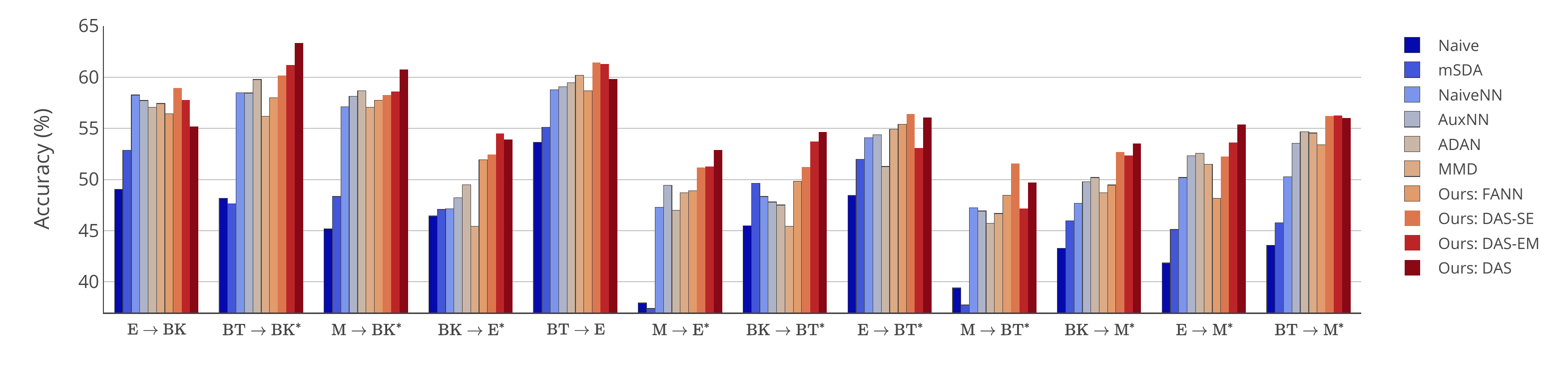}
        \caption{Accuracy on the small-scale dataset under setting 2.}
    \end{subfigure}
    
    \begin{subfigure}[b]{0.75\textwidth}
        \centering
        \includegraphics[width=\textwidth]{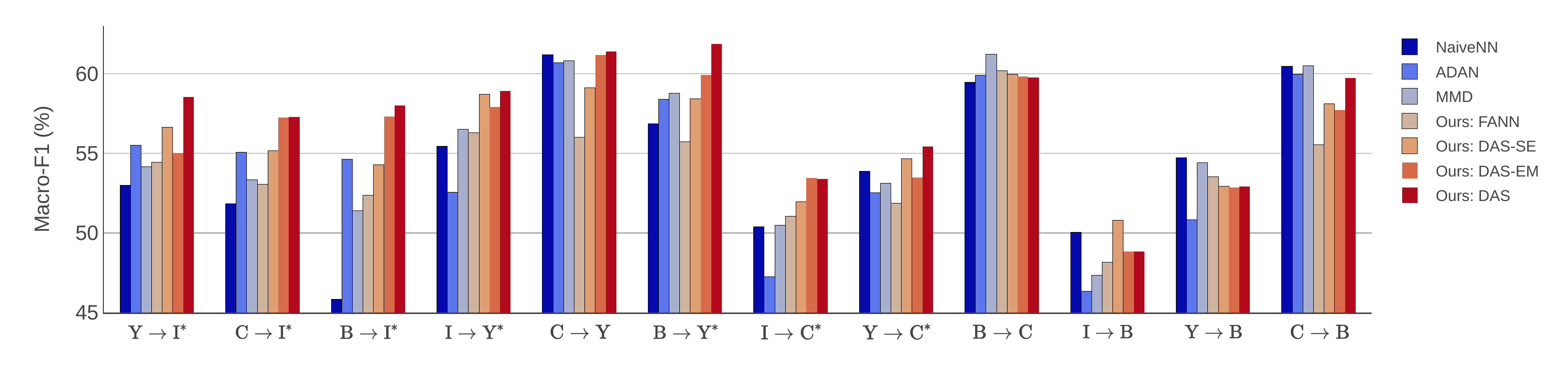}
        \caption{Macro-F1 on the large-scale dataset.}
    \end{subfigure}
    
    \caption{Performance comparison. Average results over 5 runs with random initializations are reported for each neural method. $^*$ indicates that the proposed method (either of DAS, DAS-EM, DAS-SE) is significantly better than other baselines (baseline 1-6) with $p<0.05$ based on one-tailed unpaired t-test.}
    \label{main_results}
\end{figure*}

\subsection{Training Details and Hyper-parameters}

We initialize word embeddings using the 300-dimension GloVe vectors supplied by Pennington et al., (\citeyear{Pennington:14}), which were trained on 840 billion tokens from the Common Crawl. For each pair of domains, the vocabulary consists of the top 10000 most frequent words. For words in the vocabulary but not present in the pre-trained embeddings, we randomly initialize them.

We set hyper-parameters of the CNN encoder following previous works~\cite{Kim:14,Yu:16} without specific tuning on our datasets. The window size is set to 3 and the size of the hidden layer is set to 300. The nonlinear activation function is Relu. For regularization, we also follow their settings and employ dropout with probability set to 0.5 on $\bm{\xi}_i$ before feeding it to the output layer $\mathcal{F}$, and constrain the $l_2$-norm of the weight vector $\theta_f$, setting its max norm to 3.

On the small-scale datasets and the Aamzon benchmark, $\lambda_1$ and $\lambda_2$ are set to 200 and 1, respectively, tuned on the development set of task E$\rightarrow$BK under setting 1. On the large-scale datasets, $\lambda_1$ and $\lambda_2$ are set to 500 and 0.2, respectively, tuned on I$\rightarrow$Y. We use a Gaussian curve $w(t)=\exp[-5(1-\frac{t}{t_{max}})^2]\lambda_3$ to ramp up the weight of the bootstrapping loss $\Omega$ from $0$ to $\lambda_3$, where $t_{max}$ denotes the maximum number of training epochs. We train 30 epochs for all experiments. We set $\lambda_3$ to 3 and $\alpha$ to 0.5 for all experiments. 

The batch size is set to 50 on the small-scale datasets and the Amazon benchmark. We increase the batch size to 250 on the large-scale datasets to reduce the number of iterations. RMSProp optimizer with learning rate set to 0.0005 is used for all experiments.

\subsection{Models for Comparison}
We compare with the following baselines:

(1) \textbf{Naive}: A non-domain-adaptive baseline with bag-of-words representations and SVM classifier trained on the source domain.

(2) \textbf{mSDA} \cite{chen:12}: This is the state-of-the-art method based on discrete input features. 
Top 1000 bag-of-words features are kept as pivot features. We set the number of stacked layers to 3 and the corruption probability to 0.5.

(3) \textbf{NaiveNN}: This is a non-domain-adaptive CNN trained on source domain, which is a variant of our model by setting $\lambda_1$, $\lambda_2$, and $\lambda_3$ to zeros.

(4) \textbf{AuxNN}~\cite{Yu:16}: This is a neural model that exploits auxiliary tasks, which has achieved state-of-the-art results on cross-domain sentiment classification. 
The sentence encoder used in this model is the same as ours. 

(5) \textbf{ADAN}~\cite{Chen:17}: This method exploits adversarial training to reduce representation difference between domains. The original paper uses a simple feedforward network as encoder. For fair comparison, we replace it with our CNN-based encoder. 
We train 5 iterations on the discriminator per iteration on the encoder and sentiment classifier as suggested in their paper.

(6) \textbf{MMD}: MMD has been widely used for minimizing domain discrepancy on images. In those works~\cite{tzeng:14,Long:17}, variants of deep CNNs are used for encoding images and the MMDs of multiple layers are jointly minimized. In NLP, adding more layers of CNNs may not be very helpful and thus those models from image-related tasks can not be directly applied to our problem. To compare with MMD-based method, we train a model that jointly minimize the classification loss $L$ on the source domain and MMD between $\{\bm{\xi_i^{(s)}}|_{i=1}^{n_s}\}$ and $\{\bm{\xi_i^{(t)}}|_{i=1}^{n_t}\}$. For computing MMD, we use a Gaussian RBF which is a common choice for characteristic kernel.

In addition to the above baselines, we also show results of different variants of our model. \textbf{DAS} as shown in Algorithm \ref{algo} denotes our full model. \textbf{DAS-EM} denotes the model with only entropy minimization for semi-supervised learning (set $\lambda_3=0$). \textbf{DAS-SE} denotes the model with only self-ensemble bootstrapping for semi-supervised learning (set $\lambda_2=0$). \textbf{FANN} (feature-adaptation neural network) denotes the model without semi-supervised learning performed (set both $\lambda_2$ and $\lambda_3$ to zeros).

\begin{figure*}[t!]
    \centering
    \begin{subfigure}[b]{0.25\textwidth}
        \centering
        \includegraphics[width=\textwidth]{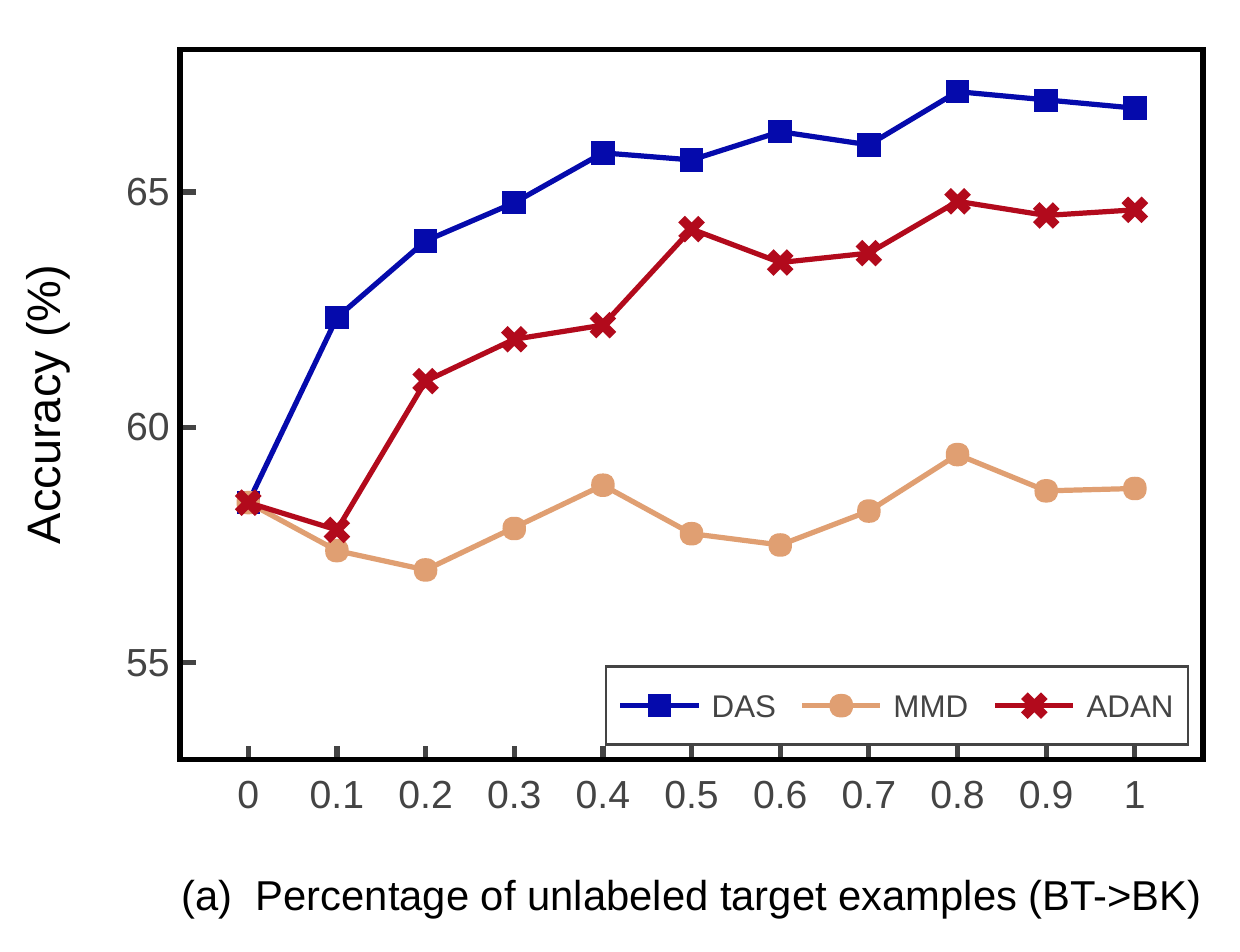}
    \end{subfigure}
    \begin{subfigure}[b]{0.25\textwidth}
        \centering
        \includegraphics[width=\textwidth]{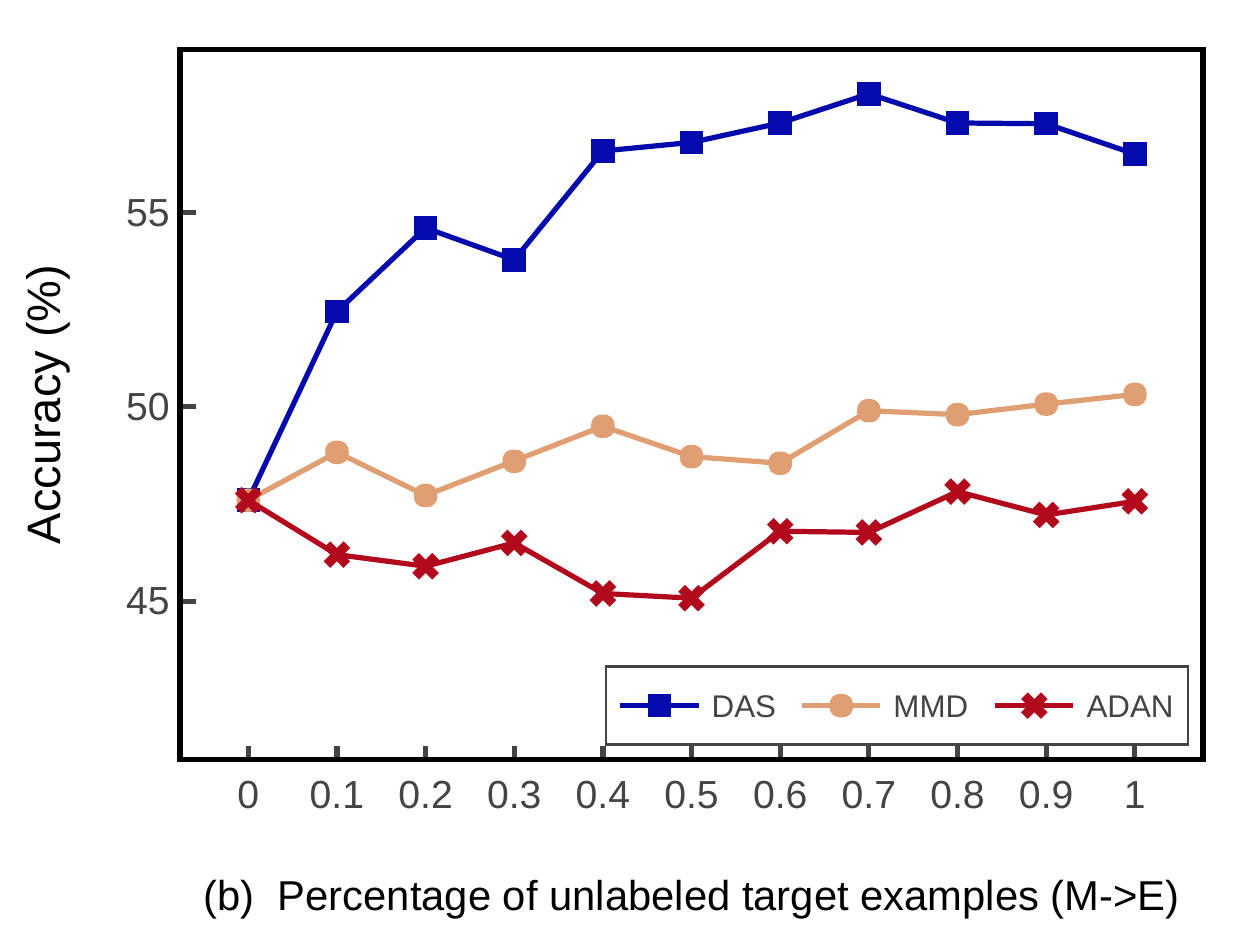}
    \end{subfigure}
    \begin{subfigure}[b]{0.25\textwidth}
        \centering
        \includegraphics[width=\textwidth]{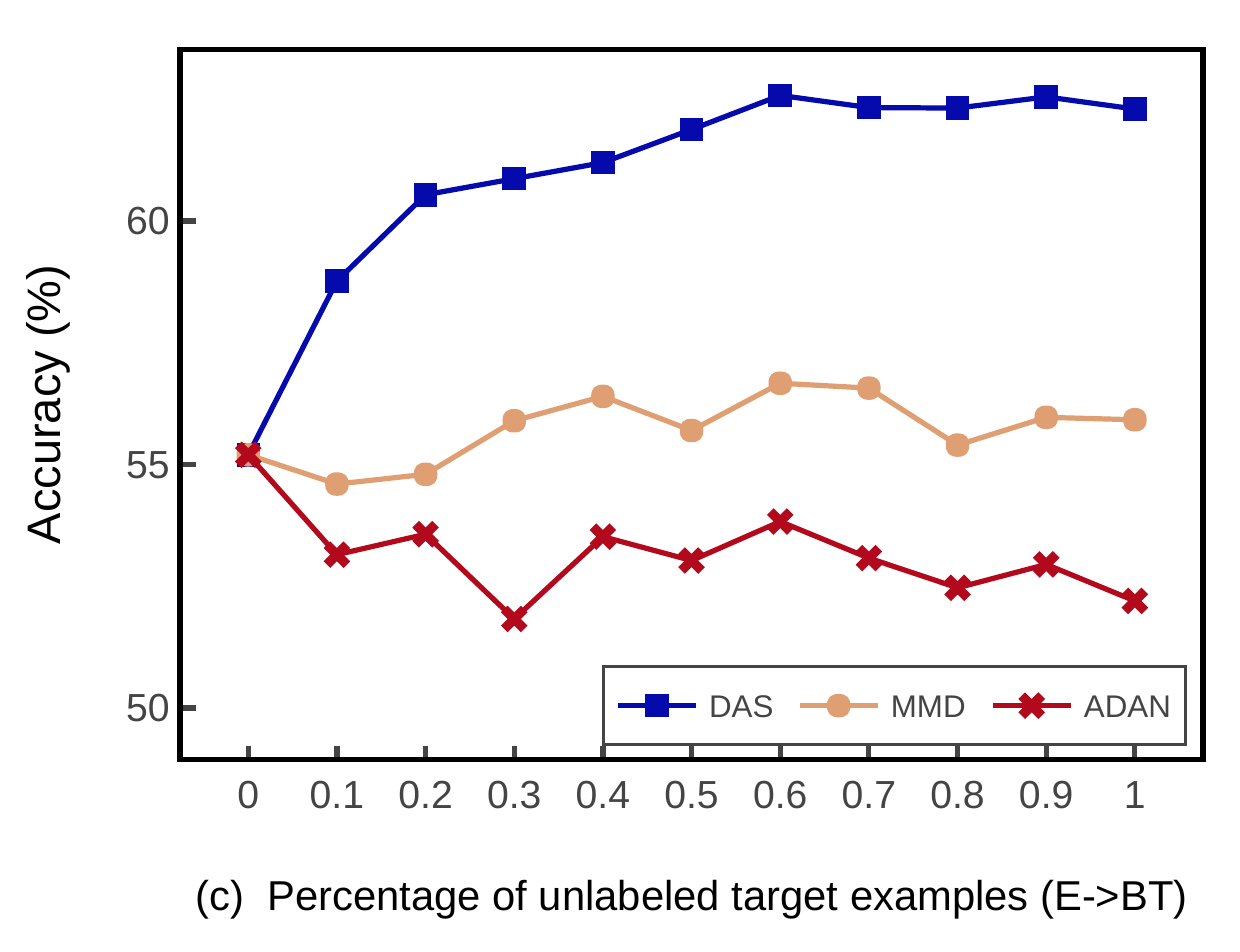}
    \end{subfigure}
    \caption{Accuracy vs. percentage of unlabeled target training examples.}
    \label{unlabel_target}
\end{figure*}

\begin{figure*}[t]
    \centering
    \begin{subfigure}[b]{0.25\textwidth}
        \centering
        \includegraphics[width=\textwidth]{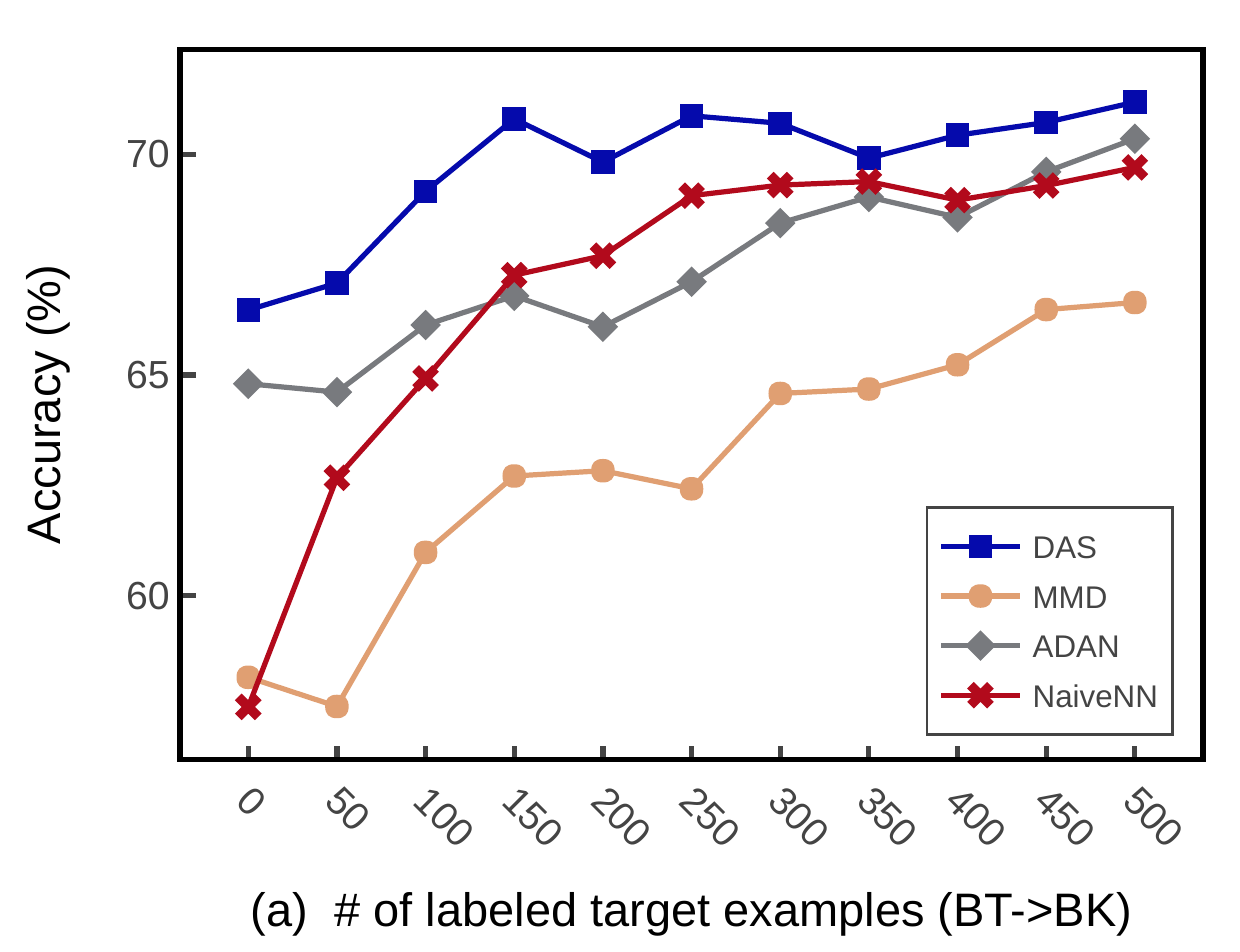}
    \end{subfigure}
    \begin{subfigure}[b]{0.25\textwidth}
        \centering
        \includegraphics[width=\textwidth]{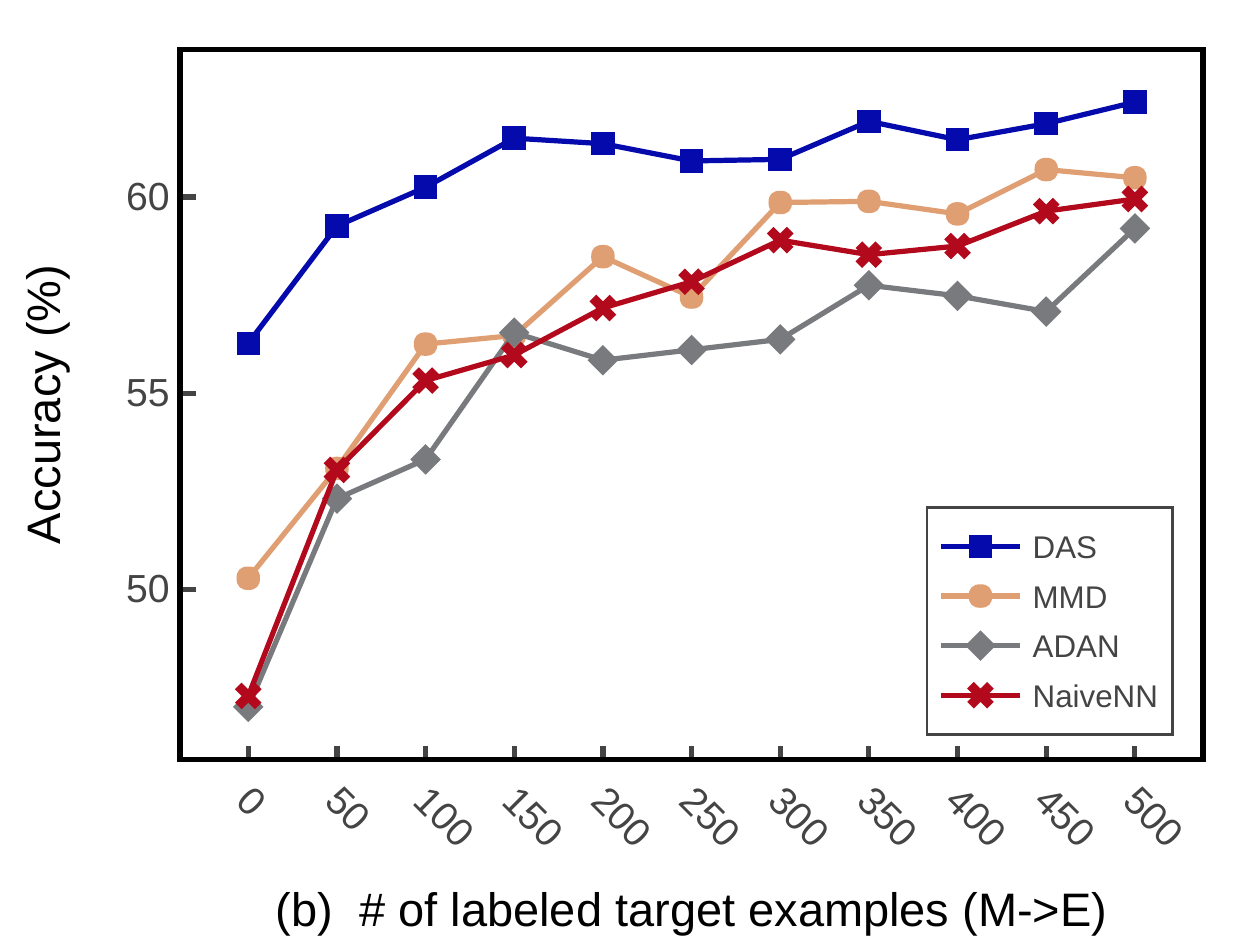}
    \end{subfigure}
    \begin{subfigure}[b]{0.25\textwidth}
        \centering
        \includegraphics[width=\textwidth]{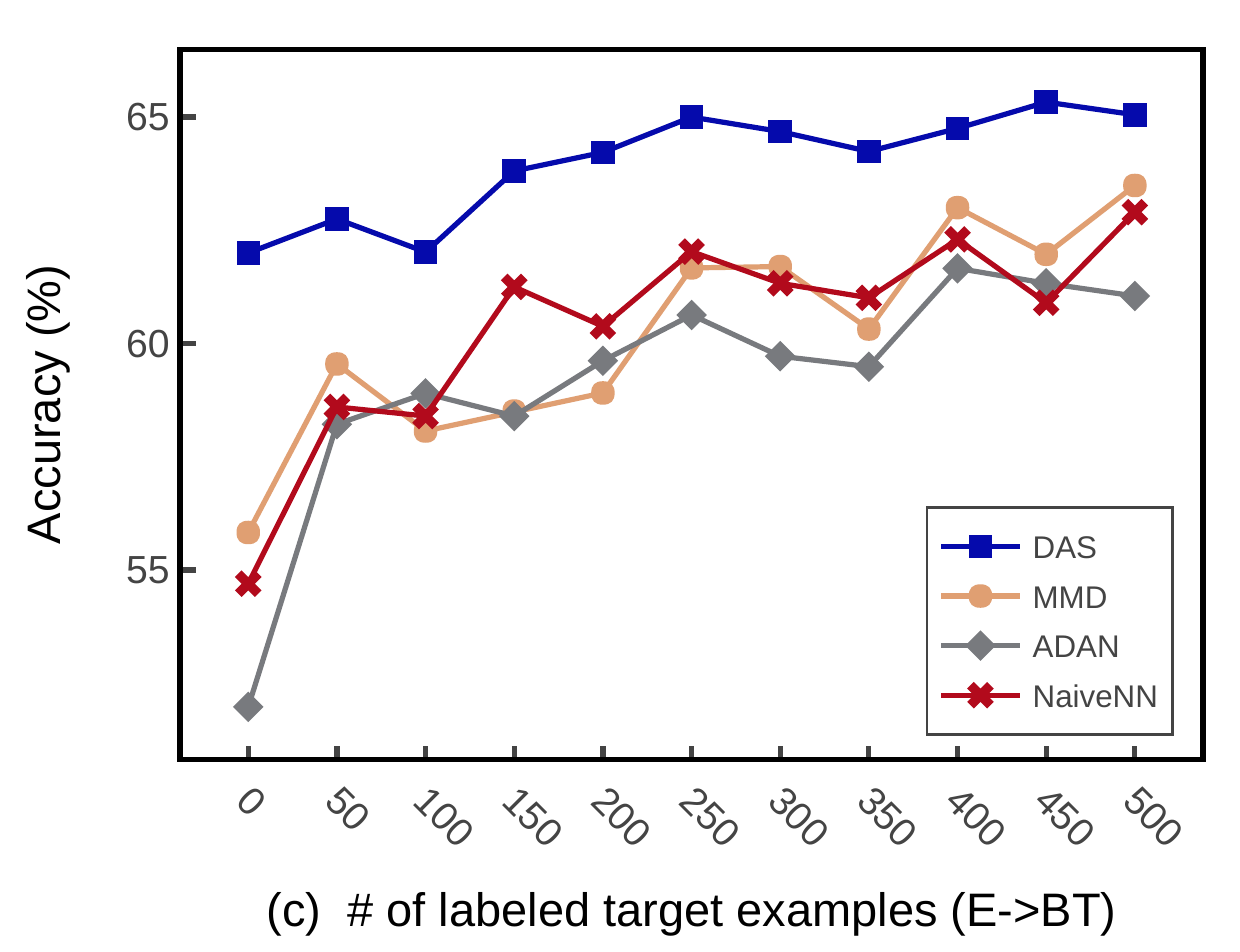}
    \end{subfigure}
    \caption{Accuracy vs. number of labeled target training examples.}
    \label{label_target}
\end{figure*}

\subsection{Main Results}
Figure~\ref{main_results}\footnote{We exclude results of Naive, mSDA and AuxNN on the large-scale dataset. Both Naive and mSDA have difficulties to scale up to the large dataset. AuxNN relies on manually selecting positive and negative pivots before training.} shows the comparison of adaptation results (see Appendix~\ref{append_numerical} for the exact numerical numbers). We report classification accuracy on the small-scale dataset. For the large-scale dataset, macro-F1 is instead used since the label distribution in the test set is extremely unbalanced. Key observations are summarized as follows. (1) Both DAS-EM and DAS-SE perform better in most cases compared with ADAN, MDD, and FANN, in which only feature adaptation is performed. This demonstrates the effectiveness of the proposed domain adaptive semi-supervised learning framework. DAS-EM is more effective than DAS-SE in most cases, and the full model DAS with both techniques jointly employed overall has the best performance. (2) When comparing the two settings on the small-scale dataset, all domain-adaptive methods\footnote{Results of Naive and NaiveNN do not change under both settings as they are only trained on the source domain.} generally perform better under setting 1. In setting 1, the target examples are balanced in classes, which can provide more diverse opinion-related features. However, when considering unsupervised domain adaptation, we should not presume the label distribution of the unlabeled data. Thus, it is necessary to conduct experiments 
using datasets that reflect real-life sentiment distribution as what we did on setting2 and the large-scale dataset. Unfortunately, this is ignored by most of previous works. (3) Word-embeddings are very helpful, as we can see even NaiveNN can substantially outperform mSDA on most tasks.

To see the effect of semi-supervised learning alone, we also conduct experiments by setting $\lambda_1=0$ to eliminate the effect of feature adaptation. Both entropy minimization and bootstrapping perform very badly in this setting. Entropy minimization gives almost random predictions with accuracy below 0.4, and the results of bootstrapping are also much lower compared to NaiveNN. This suggests that the feature adaptation component is essential. Without it, the learned target representations are less meaningful and discriminative. Applying semi-supervised learning in this case is likely to worsen the results.

\renewcommand{\arraystretch}{1.1}
\begin{table*}[t]
\begin{subtable}{1\textwidth}
\centering
\scalebox{0.5}{
\begin{tabular}{lllll}
\toprule
best-value-at&highly-recommend-!&nars-are-amazing&beauty-store-suggested&since-i-love\\
good-value-at&highly-advise-!&ulta-are-fantastic&durable-machine-and&years-i-love\\
perfect-product-for&gogeous-absolutely-perfect&length-are-so&perfect-length-and&bonus-i-love\\
great-product-at&love-love-love&expected-in-perfect&great-store-on&appearance-i-love\\
amazing-product-$*$&highly-recommend-for&setting-works-perfect&beauty-store-for&relaxing-i-love\\
\bottomrule
\end{tabular}
}
\caption{NaiveNN}\label{Naive visualization small}
\end{subtable}
\medskip

\begin{subtable}{1\textwidth}
\centering
\scalebox{0.5}{
\begin{tabular}{lllll}
\toprule
prices-my-favorite&so-nicely-!&purchase-thanks-!&feel-wonderfully-clean&are-really-cleaning\\
brands-my-favorite&more-affordable-price&buy-again-!&on-nicely-builds&washing-and-cleaning\\
very-great-stores&shampoo-a-perfect&without-hesitation-!&polish-easy-and&really-good-shampoo\\
great-bottle-also&an-excellent-value&buy-this-!&felt-cleanser-than&deeply-cleans-my\\
scent-pleasantly-floral&really-enjoy-it&discount-too-!&honestly-perfect-it&totally-moisturize-our\\
\bottomrule
\end{tabular}
}
\caption{FANN}\label{FANN visualization small}
\end{subtable}
\medskip

\begin{subtable}{1\textwidth}
\centering
\scalebox{0.5}{
\begin{tabular}{lllll}
\toprule
bath-'s-wonderful&love-fruity-sweet&feeling-smooth-radiant&cleans-thoroughly-*&excellent-everyday-lotion\\
all-pretty-affordable&absorb-really-nicely&love-lavender-scented&loving-this-soap&affordable-cleans-nicely\\
it-delivers-fabulous&shower-lather-wonderfully&am-very-grateful&bed-of-love&fantastic-base-coat\\
and-blends-nicely&*-smells-fantastic&love-fruity-fragrances&shower-!-*&nice-gentle-scrub\\
heats-quickly-love&and-clean-excellent&perfect-beautiful-shimmer&radiant-daily-moisturizer&surprisingly-safe-on\\
\bottomrule
\end{tabular}
}
\caption{DAS}\label{This work visualization small}
\end{subtable}

\caption{Comparison of the top trigrams (each column) from the target domain (beauty) captured by the 5 most positive-sentiment-related CNN filters learned on E$\rightarrow$BT. $*$ denotes a padding.}\label{visualization small}
\end{table*}

\subsection{Further Analysis}
In Figure~\ref{unlabel_target}, we show the change of accuracy with respect to the percentage of unlabeled data used for training on three particular problems under setting 1. The value at $x=0$ denotes the accuracies of NaiveNN which does not utilize any target data. For DAS, we observe a nonlinear increasing trend where the accuracy quickly improves at the beginning, and then gradually stabilizes. For other methods, this trend is less obvious, and adding more unlabeled data sometimes even worsen the results. This finding again suggests that the proposed approach can better exploit the information from unlabeled data.

We also conduct experiments under a setting with a small number of labeled target examples available. Figure~\ref{label_target} shows the change of accuracy with respect to the number of labeled target examples added for training. We can observe that DAS is still more effective under this setting, while the performance differences to other methods gradually decrease with the increasing number of labeled target examples.

\subsection{CNN Filter Analysis}\label{filter analysis}

In this subsection, we aim to better understand DAS by analyzing sentiment-related CNN filters. To do that, 1) we first select a list of the most related CNN filters for predicting each sentiment label (positive, negative neutral). Those filters can be identified according to the learned weights $\theta_f$ of the output layer $\mathcal{F}$. Higher weight indicates stronger relatedness. 2) Recall that in our implementation, each CNN filter has a window size of 3 with Relu activation. We can thus represent each selected filter as a ranked list of trigrams with highest activation values. 

We analyze the CNN filters learned by NaiveNN, FANN and DAS respectively on task E$\rightarrow$BT under setting 1. We focus on E$\rightarrow$BT for study because electronics and beauty are very different domains and each of them has a diverse set of domain-specific sentiment expressions. For each method, we identify the top 10 most related filters for each sentiment label, and extract the top trigrams of each selected filter on both source and target domains.
Since labeled source examples are used for training, we find the filters learned by the three methods capture similar expressions on the source domain, containing both domain-invariant and domain-specific trigrams. On the target domain, DAS captures more target-specific expressions compared to the other two methods. Due to space limitation, we only present a small subset of positive-sentiment-related filters in Table~\ref{visualization small}. The complete results are provided in Appendix~\ref{append_cnn}.
From Table~\ref{visualization small}, we can observe that the filters learned by NaiveNN are almost unable to capture target-specific sentiment expressions, while FANN is able to capture limited target-specific words such as ``clean'' and ``scent''. The filters learned by DAS are more domain-adaptive, capturing diverse sentiment expressions in the target domain. 

\section{Conclusion}
In this work, we propose DAS, a novel framework that jointly performs feature adaptation and semi-supervised learning. We have demonstrated through multiple experiments that DAS can better leverage unlabeled data, and achieve substantial improvements over baseline methods. We have also shown that feature adaptation is an essential component, without which, semi-supervised learning is not able to function properly. The proposed framework could be potentially adapted to other domain adaptation tasks, which is the focus of our future studies.

\bibliography{emnlp2018}
\bibliographystyle{acl_natbib_nourl}

\clearpage

\appendix
\section{Results on Amazon Benchmark}\label{append_amazon}
Most previous works~\cite{blitzer:07,pan:10,glorot:11,chen:12,zhou:16} carried out experiments on the Amazon benchmark released by Blitzer et al. (\citeyear{blitzer:07}). The dataset contains 4 different domains: Book (B), DVDs (D), Electronics (E), and Kitchen (K). Following their experimental settings, we consider the binary classification task to predict whether a review is positive or negative on the target domain. Each domain consists of 1000 positive and 1000 negative reviews respectively. We also allow 4000 unlabeled reviews to be used for both the source and the target domains, of which the positive and negative reviews are balanced as well, following the settings in previous works. We construct 12 cross-domain sentiment classification tasks and split the labeled data in each domain into a training set of 1600 reviews and a test set of 400 reviews. The classifier is trained on the training set of the source domain and is evaluated on the test set of the target domain. The comparison results are shown in Table~\ref{amazon_benchmark}.

\begin{table*}[t]
\centering
\small
\begin{tabular}{cc|ccccccc|ccc}
\toprule
S &T &Naive &mSDA &NaiveNN &AuxNN &ADAN &MMD &FANN &DAS-EM &DAS-SE &DAS \\\hline\hline
D&B&75.20&78.50&81.10&80.80&81.70&81.05&80.30&82.00$^*$&\bf{82.10}$^*$&82.05$^*$\\
E&B&68.85&76.15&77.95&78.00&78.55&78.65&77.25&\bf{80.25}$^*$&77.75&80.00$^*$\\
K&B&70.00&75.65&77.75&77.85&79.25&79.70&79.20&79.95&79.60&\bf{80.05}$^*$\\\hline
B&D&77.15&80.60&80.80&81.75&82.30&82.00&81.65&82.65&82.35&\bf{82.75}$^*$\\
E&D&69.50&76.30&77.00&80.65&79.70&80.10&79.55&\bf{81.40}$^*$&79.75&80.15\\
K&D&71.40&76.05&79.35&78.90&80.45&79.35&76.90&81.65$^*$&\bf{82.15}$^*$&81.40$^*$\\\hline
B&E&72.15&75.55&76.20&76.40&77.60&76.45&76.75&80.25$^*$&75.80&\bf{81.15}$^*$\\
D&E&71.65&76.00&76.60&77.55&79.70&80.20&79.25&81.40$^*$&80.05&\bf{81.55}$^*$\\
K&E&79.75&84.20&84.85&84.05&\bf{86.85}&85.75&85.60&85.70&85.95&85.80\\\hline
B&K&73.50&75.95&77.40&78.10&76.10&75.20&77.55&81.55$^*$&79.45$^*$&\bf{82.25}$^*$\\
D&K&72.00&76.30&78.55&80.05&77.35&79.70&78.00&80.80&79.50&\bf{81.50}$^*$\\
E&K&82.80&84.45&\bf{84.95}&84.15&83.95&81.75&83.85&84.50&83.80&84.85\\\hline\hline
\multicolumn{2}{c|}{Average}&73.66&77.98&79.38&79.85&80.29&80.00&79.65&81.84&80.68&\bf{81.96}\\\bottomrule
\end{tabular}
\caption{Accuracies on the Amazon benchmark. Average results over 5 runs with random initializations are reported for each neural method. $^*$ indicates that the proposed method (DAS-EM, DAS-SE, DAS) is significantly better than other baselines with $p<0.05$ based on one-tailed unpaired t-test.}\label{amazon_benchmark}
\end{table*}

\section{Numerical Results of Figure~\ref{main_results}}\label{append_numerical}

Due to space limitation, we only show results in figures in the paper. All numerical numbers used for plotting Figure~\ref{main_results} are presented in Table~\ref{numerical_results}. We can observe that DAS-EM, DAS-SE, and DAS all achieve substantial improvements over baseline methods under different settings. 

\begin{table*}[t]
\begin{subtable}{1\textwidth}
\centering
\small
\begin{tabular}{cc|ccccccc|ccc}
\toprule
S &T &Naive &mSDA &NaiveNN &AuxNN &ADAN &MMD &FANN &DAS-EM &DAS-SE &DAS \\\hline\hline
E&BK   &49.07&55.13&58.26&60.62&63.32&60.38&59.59&66.48$^*$&62.37&\bf{67.12}$^*$\\
BT&BK  &48.17&53.53&58.48&59.86&65.62&59.66&59.28&\bf{66.78}$^*$&61.17&66.53\\
M&BK   &45.20&49.22&57.10&60.43&62.87&60.20&57.65&69.63$^*$&65.24$^*$&\bf{70.31}$^*$\\
BK&E   &46.43&48.22&47.15&48.45&47.42&53.32&51.27&58.59$^*$&55.15$^*$&\bf{58.73}$^*$\\
BT&E   &53.63&57.32&58.77&60.98&63.13&60.53&60.62&65.71$^*$&61.78&\bf{66.14}$^*$\\
M&E    &37.93&38.13&47.28&49.60&46.57&51.55&47.23&\bf{55.88}$^*$&53.22$^*$&55.78$^*$\\
BK&BT  &45.57&50.77&48.35&48.67&46.14&49.48&50.24&49.49&\bf{54.23}$^*$&51.30$^*$\\
E&BT   &48.43&53.13&54.07&55.58&50.98&54.83&56.78&\bf{61.53}$^*$&59.52$^*$&60.76$^*$\\
M&BT   &39.42&39.37&47.23&48.65&44.26&48.35&48.89&47.65&\bf{50.67}$^*$&50.66$^*$\\
BK&M   &43.32&47.88&47.67&48.87&51.10&53.04&52.35&55.47$^*$&55.13$^*$&\bf{55.98}$^*$\\
E&M    &41.83&47.88&50.21&51.19&50.23&51.81&52.14&58.28$^*$&55.60$^*$&\bf{59.06}$^*$\\
BT&M   &43.55&49.62&50.27&53.11&55.35&54.43&53.84&\bf{60.95}$^*$&56.90$^*$&60.5$^*$\\\hline
\multicolumn{2}{c|}{Average}&45.21&49.18&52.07&53.84&53.92&54.80&54.15&59.74&57.58&\bf{60.24}\\
\bottomrule
\end{tabular}
\caption{Accuracy on the small-scale dataset under setting 1}
\end{subtable}
\bigskip

\begin{subtable}{1\textwidth}
\centering
\small
\begin{tabular}{cc|ccccccc|ccc}
\toprule
S &T &Naive &mSDA &NaiveNN &AuxNN &ADAN &MMD &FANN &DAS-EM &DAS-SE &DAS \\\hline\hline
E&BK   &49.07&52.88&58.26&57.72&57.07&57.43&56.43&57.78&\bf{58.93}&55.20\\
BT&BK  &48.17&47.65&58.48&58.46&59.78&56.17&57.98&61.17$^*$&60.17$^*$&\bf{63.32}$^*$\\
M&BK   &45.20&48.33&57.10&58.15&58.67&57.08&57.75&58.62&58.25&\bf{60.77}$^*$\\
BK&E   &46.43&47.07&47.15&48.22&49.48&45.42&51.95&\bf{54.51}$^*$&52.47$^*$&53.92$^*$\\
BT&E   &53.63&55.12&58.77&59.08&59.45&60.24&58.67&61.27&\bf{61.42}&59.83\\
M&E    &37.93&37.40&47.28&49.43&47.00&48.72&48.92&51.28$^*$&51.18$^*$&\bf{52.88}$^*$\\
BK&BT  &45.57&49.63&48.35&47.80&47.52&45.43&49.83&53.72$^*$&51.23$^*$&\bf{54.67}$^*$\\
E&BT   &48.43&51.98&54.07&54.37&51.28&54.92&55.42&53.10&\bf{56.43}$^*$&56.05$^*$\\
M&BT   &39.43&37.73&47.23&46.92&45.73&46.68&48.48&47.18&\bf{51.57}$^*$&49.73$^*$\\
BK&M   &43.32&45.97&47.67&48.79&50.20&48.76&49.47&52.37$^*$&52.68$^*$&\bf{53.52}$^*$\\
E&M    &41.83&45.12&50.21&52.31&52.57&51.50&48.18&53.63$^*$&52.25&\bf{55.38}$^*$\\
BT&M   &43.55&45.78&50.27&53.55&54.68&54.55&53.41&\bf{56.24}$^*$&56.23$^*$&56.02$^*$\\\hline
\multicolumn{2}{c|}{Average}&45.21&47.06&52.07&52.98&52.79&52.23&53.04&55.07&55.23&\bf{55.94}\\
\bottomrule
\end{tabular}
\caption{Accuracy on the small-scale dataset under setting 2}
\end{subtable}
\bigskip

\begin{subtable}{1\textwidth}
\centering
\small
\begin{tabular}{cc|cccc|ccc}
\toprule
S &T &NaiveNN &ADAN &MMD &FANN &DAS-EM &DAS-SE &DAS \\\hline\hline
Y&I&53.01&55.52&54.16&54.46&55.04&56.66$^*$&\bf{58.54}$^*$\\
C&I&51.84&55.07&53.35&53.07&57.27$^*$&55.18&\bf{57.28}$^*$\\
B&I&45.85&54.64&51.40&52.39&57.31$^*$&54.30&\bf{58.02}$^*$\\
I&Y&55.46&52.57&56.52&56.30&57.92$^*$&58.72$^*$&\bf{58.92}$^*$\\
C&Y&61.22&60.70&60.81&56.02&61.17&59.14&\bf{61.39}\\
B&Y&56.86&58.42&58.77&55.72&59.94$^*$&58.43&\bf{61.87}$^*$\\
I&C&50.38&47.27&50.49&51.04&\bf{53.46}$^*$&51.97$^*$&53.38$^*$\\
Y&C&53.87&52.53&53.12&51.86&53.48&\bf{54.67}$^*$&\bf{55.44}$^*$\\
B&C&59.48&59.91&\bf{61.23}&60.19&59.84&59.98&59.76\\
I&B&50.05&46.34&47.35&48.17&48.84&\bf{50.81}&48.84\\
Y&B&\bf{54.73}&50.82&54.43&53.54&52.87&52.95&52.91\\
C&B&60.47&59.99&\bf{60.52}&55.56&57.74&58.12&59.75\\\hline
\multicolumn{2}{c|}{Average}&54.43&54.48&55.18&54.02&56.24&55.91&\bf{57.18}\\
\bottomrule
\end{tabular}
\caption{Macro-F1 scores on the large-scale dataset}
\end{subtable}

\caption{Performance comparison. Average results over 5 runs with random initializations are reported for each neural method. $^*$ indicates that the proposed method (DAS, DAS-EM, DAS-SE) is significantly better than other baselines with $p<0.05$ based on one-tailed unpaired t-test.}
\label{numerical_results}
\end{table*}

\section{CNN Filter Analysis Full Results}\label{append_cnn}

As mentioned in Section \ref{filter analysis}, we conduct CNN filter analysis on NaiveNN, FANN, and DAS. For each method, we identify the top 10 most related filters for positive, negative, neutral sentiment labels respectively, and then represent each selected filter as a ranked list of trigrams with the highest activation values on it. Table \ref{pos_related}, \ref{neg_related}, \ref{neu_related} in the following pages illustrate the trigrams from the target domain (beauty) captured by the selected filters learned on E$\rightarrow$BT for each method.

We can observe that compared to NaiveNN and FANN, DAS is able to capture a more diverse set of relevant sentiment expressions on the target domain for each sentiment label. This observation is consistent with our motivation. Since NaiveNN, FANN and other baseline methods solely train the sentiment classifier on the source domain, the learned encoder is not able to produce discriminative features on the target domain. DAS addresses this problem by refining the classifier on the target domain with semi-supervised learning, and the overall objective forces the encoder to learn feature representations that are not only domain-invariant but also discriminative on both domains.

\renewcommand{\arraystretch}{1.1}
\begin{table*}[t]
\begin{subtable}{1\textwidth}
\centering
\scalebox{0.7}{
\begin{tabular}{lllll}
\toprule
1&2&3&4&5\\
best-value-at&highly-recommend-!&nars-are-amazing&beauty-store-suggested&since-i-love\\
good-value-at&highly-advise-!&ulta-are-fantastic&durable-machine-and&years-i-love\\
perfect-product-for&gogeous-absolutely-perfect&length-are-so&perfect-length-and&bonus-i-love\\
great-product-at&love-love-love&expected-in-perfect&great-store-on&appearance-i-love\\
amazing-product-$*$&highly-recommend-for&setting-works-perfect&beauty-store-for&relaxing-i-love\\
\\

6&7&8&9&10\\
store-and-am&office-setting-thanks&car-washes-!&speed-is-perfect&!-i-recommend\\
cleanser-and-am&locks-shimmering-color&price-in-stores&buttons-are-perfect&!-i-highly\\
olay-and-am&dirty-blonde-color&products-are-priced&unit-is-superb&shower-i-slather\\
daily-and-need&victoria-secrets-gorgeous&car-and-burning&spray-is-perfect&spots-i-needed\\
shower-and-noticed&dirty-pinkish-color&from-our-store&coverage-is-excellent&best-i-use\\

\bottomrule
\end{tabular}
}
\caption{NaiveNN}
\end{subtable}
\medskip

\begin{subtable}{1\textwidth}
\centering
\scalebox{0.7}{
\begin{tabular}{lllll}
\toprule
1&2&3&4&5\\
prices-my-favorite&so-nicely-!&purchase-thanks-!&feel-wonderfully-clean&are-really-cleaning\\
brands-my-favorite&more-affordable-price&buy-again-!&on-nicely-builds&washing-and-cleaning\\
very-great-stores&shampoo-a-perfect&without-hesitation-!&polish-easy-and&really-good-shampoo\\
great-bottle-also&an-excellent-value&buy-this-!&felt-cleanser-than&deeply-cleans-my\\
scent-pleasantly-floral&really-enjoy-it&discount-too-!&honestly-perfect-it&totally-moisturize-our\\
\\

6&7&8&9&10\\
shower-or-cleaning&definitely-purchase-again&more-affordable-price&absolutely-wonderful-!&felt-cleaner-than\\
water-onto-my&definitely-buy-again&a-perfect-length&perfect-for-running&flat-iron-through\\
bleach-your-towels&perfect-for-my&an-exceptional-value&concealer-for-my&rubbed-grease-on\\
pump-onto-my&definitely-order-again&'ve-enjoyed-it&moisturizing-for-my&deeply-cleans-my\\
water-great-for&super-happy-to&pretty-decent-layer&super-glue-even&being-cleaner-after\\

\bottomrule
\end{tabular}
}
\caption{FANN}
\end{subtable}
\medskip

\begin{subtable}{1\textwidth}
\centering
\scalebox{0.7}{
\begin{tabular}{lllll}
\toprule
1&2&3&4&5\\
bath-'s-wonderful&love-fruity-sweet&feeling-smooth-radiant&cleans-thoroughly-*&excellent-everyday-lotion\\
all-pretty-affordable&absorb-really-nicely&love-lavender-scented&loving-this-soap&affordable-cleans-nicely\\
it-delivers-fabulous&shower-lather-wonderfully&am-very-grateful&bed-of-love&fantastic-base-coat\\
and-blends-nicely&*-smells-fantastic&love-fruity-fragrances&shower-!-*&nice-gentle-scrub\\
heats-quickly-love&and-clean-excellent&perfect-beautiful-shimmer&radiant-daily-moisturizer&surprisingly-safe-on\\
\\
6&7&8&9&10\\
shower-lather-wonderfully&highly-recommend-!&excellent-fragrance-and&its-unique-smoothing&forgeous-gragrance-mist\\
affordable-cleans-nicely&definitely-recommend-!&fantastic-for-daytime&smooth-luxurious-texture&wonderful-bedtime-scent\\
peels-great-price&love-love-!&wonderfully-moisturizing-and&'s-extremely-gentle&love-essie-polish\\
daughter-loves-this&highly-advise-!&lathers-great-cleans&'s-affordable-combination&perfect-beautiful-shimmer\\
cleans-great-smells&time-advise-!&delightful-shampoo-works&absorbs-quite-well&fantastic-coverage-hydrates\\

\bottomrule
\end{tabular}
}
\caption{DAS}
\end{subtable}

\caption{Top 5 trigrams from the target domain (beauty) captured by the top 10 most \textbf{positive-sentiment-related} CNN filters learned on E$\rightarrow$BT. $*$ denotes a padding.}\label{pos_related}
\end{table*}

\renewcommand{\arraystretch}{1.1}
\begin{table*}[t]
\begin{subtable}{1\textwidth}
\centering
\scalebox{0.7}{
\begin{tabular}{lllll}
\toprule
1&2&3&4&5\\
pads-ruined-my&simply-threw-out&hours-after-trying&junk-drawer-$*$&contacted-manufacturer-about\\
highly-disappointed-and&reviewer-pointed-out&minutes-after-rinsing&refund-time-!&minutes-not-worth\\
dryers-blew-my&extracts-broke-into&disappointed-after-trying&total-fake-wen&'ve-owned-this\\
completely-worthless-didn't&actually-threw-out&lips-after-trying&waste-your-time&hour-unless-it\\
am-disappointed-and&clips-barely-keep&dry-after-shampooing&total-fail-!&results-they-claim\\
\\

6&7&8&9&10\\
were-awful-garbage&two-failed-attempts&auto-ship-sent&refund-and-dispose&broke-don't-fix\\
what-awful-garbage&a-mistake-save&am-returning-to&refund-spend-your&sent-me-expired\\
and-utter-waste&a-definite-return&am-unable-to&wouldn't-recommend-!&wearing-false-eyelashes\\
are-absolute-garbage&a-pathetic-limp&am-pale-ghost&not-buy-dunhill&a-temporary-fix\\
piece-of-junk&a-total-disappointment&got-returned-and&not-worth-returning&a-disappointment-cheap\\

\bottomrule
\end{tabular}
}
\caption{NaiveNN}
\end{subtable}
\medskip

\begin{subtable}{1\textwidth}
\centering
\scalebox{0.7}{
\begin{tabular}{lllll}
\toprule
1&2&3&4&5\\
nasty-sunburn-lol&the-worse-mascaras&stale-very-unhappy&actually-hurts-your&a-return-label\\
bother-returning-them&it-caused-patchy&were-horrible-failures&didn't-bother-returning&stay-away-completely\\
fails-miserably-at&lifeless-disaster-enter&send-this-crap&it-hurts-your&like-bug-quit\\
minutes-auric-needs&it-fails-miserably&were-awful-garbage&didn't-exist-in&a-defective-brown\\
severely-burned-me&feel-worse-leaving&were-horribly-red&skin-horribly-after&'d-refund-the\\
\\

6&7&8&9&10\\
worse-with-exercise&not-stink-your&it-fails-miserably&got-progressively-worse&stopped-working-for\\
worse-and-after&mistake-save-your&is-ineffective-apart&gave-opposite-result&uncomfortable-i-returned\\
unable-to-return&nothing-!-by&but-horribly-unhealthy&another-epic-fail&i-am-returning\\
worse-my-face&nothing-happened-!&a-pathetic-limp&got-horribly-painful&stopped-working-shortly\\
poorly-in-step&nothing-save-your&a-worse-job&was-downright-painful&not-waterproof-makeup\\

\bottomrule
\end{tabular}
}
\caption{FANN}
\end{subtable}
\medskip

\begin{subtable}{1\textwidth}
\centering
\scalebox{0.7}{
\begin{tabular}{lllll}
\toprule
1&2&3&4&5\\
poorly-designed-product&a-refund-spend&completely-waste-of&smells-disgusting-!&burning-rubber-stench\\
defective-dryer-promising&a-refund-save&of-junk-*&smells-horribly-like&began-smelling-vomit\\
disgusting-smelling-thing&i-regret-spending&were-awful-garbage&does-not-straighten&reaction-and-wasted\\
hurts-your-scalp&just-wouldn't-spend&worthless-waste-of&'s-false-advertising&control-and-smelled\\
hurts-your-hair&looked-washed-out&throwing-money-away&a-disgusting-cheap&using-this-disgusting\\

\\
6&7&8&9&10\\
super-irritating-!&got-promptly-broke&sore-and-painful&it-caused-patchy&painful-it-hurt\\
strong-reaction-and&after-ive-washed&is-simply-irritating&layer-hydrogenated-alcohols&unnecessary-health-risks\\
really-burned-and&after-several-attempts&tight-and-uncomfortable&the-harmful-uva&uncomfortable-to-wear\\
very-pasty-and&this-stuff-stinks&drying-and-irritating&my-severe-dark&stinging-your-eyes\\
super-streaky-and&again-i-threw&goopy-and-unpleasant&a-allergic-reaction&unbearable-to-wear\\

\bottomrule
\end{tabular}
}
\caption{DAS}
\end{subtable}

\caption{Top 5 trigrams from the target domain (beauty) captured by the top 10 most \textbf{negative-sentiment-related} CNN filters learned on E$\rightarrow$BT. $*$ denotes a padding.}\label{neg_related}
\end{table*}

\renewcommand{\arraystretch}{1.1}
\begin{table*}[t]
\begin{subtable}{1\textwidth}
\centering
\scalebox{0.7}{
\begin{tabular}{lllll}
\toprule
1&2&3&4&5\\
purpose-cologne-splash&okay-cord-was&hands-feet-elbows&aggressive-in-general&but-its-okay\\
other-hanae-mori&cocamide-dea-is&been-sealed-tight&pimples-in-general&it-moisturizes-okay\\
the-mavala-peeled&coily-conditioner-is&stainless-steel-blackhead&biotin-in-general&but-moisturizes-keeps\\
avoid-hair-pulling&flaky-dandruff-is&severely-tight-chest&dimethicone-is-terrible&but-don't-expect\\
cause-rashes-stinging&quickly-cord-is&thick-nasty-callouses&but-in-general&it-lathers-ok\\
\\

6&7&8&9&10\\
pretty-damaged-from&darker-olive-complexion&doesn't-mind-pushing&producto-por-los&feeling-didn't-last\\
daughter-suffers-from&stronger-healthier-or&kinda-doesn't-its&unstuck-frownies-$*$&curls-didn't-last\\
teenager-suffers-from&natural-ingredient-however&kinda-kinky-coily&they-push-$*$&extra-uv-protection\\
tissue-damage-during&vitamin-enriched-color&okay-job-of&uva-rays-uva&garnier-fructis-curl\\
the-damage-on&natural-ingredients-$*$&intended-purpose-that&tend-to-slip&the-mavala-after\\

\bottomrule
\end{tabular}
}
\caption{NaiveNN}
\end{subtable}
\medskip

\begin{subtable}{1\textwidth}
\centering
\scalebox{0.7}{
\begin{tabular}{lllll}
\toprule
1&2&3&4&5\\
worse-and-after&maybe-a-refund&very-disappointing-waste&my-ears-are&pretty-neutral-neither\\
worse-before-improving&ok-mask-i&ok-but-clean&my-neck-line&ok-so-if\\
unable-to-return&ok-pining-it&ok-but-will&cause-unsightly-beads&ok-during-pregnancy\\
unless-your-entire&ok-try-i&ok-but-didn't&my-sporadic-line&kinda-annoying-if\\
horrible-in-execution&ok-tho-i&ok-nothing-special&your-ear-is&ok-this-seems\\
\\

6&7&8&9&10\\
uncomfortable-i-returned&sticky-lathers-and&some-fading-when&are-very-painful&its-also-convenient\\
weak-they-bend&quickly-deep-cleans&real-disappointment-the&are-less-painful&that-also-my\\
claimed-faulty-$*$&but-elegant-bottle&especially-noticeable-after&are-a-pain&that-may-make\\
suffers-from-wind&beat-the-price&progressively-worse-during&about-sum-damage&that-allows-your\\
as-defective-$*$&and-reasonably-priced&style-unfortunately-the&offered-no-pain&its-helpful-to\\

\bottomrule
\end{tabular}
}
\caption{FANN}
\end{subtable}
\medskip

\begin{subtable}{1\textwidth}
\centering
\scalebox{0.7}{
\begin{tabular}{lllll}
\toprule
1&2&3&4&5\\
'm-kinda-pale&darker-but-nope&ok-but-horrible&noticeable-i-avoid&same-result-mediocre\\
a-terrible-headache&gray-didn't-cover&ok-but-didn't&however-i-lean&it-caused-patchy\\
but-kinda-annoying&makeup-doesn't-sweat&okay-but-doesn't&but-otherwise-ok&doesn't-cause-flare\\
'm-kinda-mad&dark-spots-around&okay-however-it&but-im-deciding&the-harmful-uva\\
i-kinda-stopped&moist-but-thats&unfortunately-straight&however-i-prefer&rather-unpleasant-smell\\

\\
6&7&8&9&10\\
kinda-annoying-if&brutal-winter-however&higher-rating-because&nothing-for-odor&but-darker-*\\
pretty-bad-breakage&summer-color-however&slight-burnt-rubber&kinda-recommend-this&slightly-darker-shade\\
my-slight-discoloration&beige-shade-however&noticeable-tan-since&not-recommend-if&somewhat-pale-affect\\
smells-kinda-bad&is-okay-however&somewhat-pale-affect&noticeable-but-non&but-somewhat-heavy\\
look-kinda-crappy&bit-greasy-however&kinda-pale-so&nothing-special-moderate&bit-dull-heavy\\

\bottomrule
\end{tabular}
}
\caption{DAS}
\end{subtable}

\caption{Top 5 trigrams from the target domain (beauty) captured by the top 10 most \textbf{neutral-sentiment-related} CNN filters learned on E$\rightarrow$BT. $*$ denotes a padding.}\label{neu_related}
\end{table*}

\end{document}